\documentclass[conference]{IEEEtran}
\IEEEoverridecommandlockouts
\usepackage{cite}
\usepackage{amsmath,amssymb,amsfonts}
\usepackage{algorithmic}
\usepackage{graphicx}
\usepackage{textcomp}
\usepackage{xcolor}
\usepackage{subcaption}
\usepackage{multicol}
\usepackage{multirow}
\usepackage{url}
\usepackage[ruled,vlined]{algorithm2e}

\def\BibTeX{{\rm B\kern-.05em{\sc i\kern-.025em b}\kern-.08em
    T\kern-.1667em\lower.7ex\hbox{E}\kern-.125emX}}
\begin{document}

\title{Contrastive Learning with Cross-Modal Knowledge Mining for Multimodal Human Activity Recognition\\
\thanks{This work has been partially funded by the European Union’s Horizon2020 project: PeRsOnalized Integrated CARE Solution for Elderly facing several short or long term conditions and enabling a better quality of LIFE (Procare4Life), under Grant Agreement N.875221.}
}

\author{\IEEEauthorblockN{Razvan Brinzea}
\IEEEauthorblockA{
\textit{Maastricht University}\\
Maastricht, Netherlands \\
r.brinzea@student.maastrichtuniversity.nl
}
\and
\IEEEauthorblockN{Bulat Khaertdinov}
\IEEEauthorblockA{
\textit{Maastricht University}\\
Maastricht, Netherlands \\
b.khaertdinov@maastrichtuniversity.nl
}
\and
\IEEEauthorblockN{Stylianos Asteriadis}
\IEEEauthorblockA{
\textit{Maastricht University}\\
Maastricht, Netherlands \\
stelios.asteriadis@maastrichtuniversity.nl
}}

\maketitle

\begin{abstract}
Human Activity Recognition is a field of research where input data can take many forms. Each of the possible input modalities describes human behaviour in a different way, and each has its own strengths and weaknesses. We explore the hypothesis that leveraging multiple modalities can lead to better recognition. Since manual annotation of input data is expensive and time-consuming, the emphasis is made on self-supervised methods which can learn useful feature representations without any ground truth labels. We extend a number of recent contrastive self-supervised approaches for the task of Human Activity Recognition, leveraging inertial and skeleton data. Furthermore, we propose a flexible, general-purpose framework for performing multimodal self-supervised learning, named Contrastive Multiview Coding with Cross-Modal Knowledge Mining (CMC-CMKM). This framework exploits modality-specific knowledge in order to mitigate the limitations of typical self-supervised frameworks. The extensive experiments on two widely-used datasets demonstrate that the suggested framework significantly outperforms contrastive unimodal and multimodal baselines on different scenarios, including fully-supervised fine-tuning, activity retrieval and semi-supervised learning. Furthermore, it shows performance competitive even compared to supervised methods.
\end{abstract}


\section{Introduction}
Human Activity Recognition (HAR) is a joint area of research in the fields of Human-Centered Computing and Human-Computer Interaction, with practical applications in many areas, such as smart homes \cite{Mehr_2019_smarthome, Yegang_2019_smarthome}, health monitoring \cite{Panwar_2019}, manufacturing automation \cite{Grzescick_2017} and sport analytics \cite{Vleugels_2021}. 

The modalities which can be used for HAR include but are not limited to RGB-D streams, skeleton data, wearable sensor data (or inertial data). Different techniques can be employed for HAR depending on the type of the input data, but each modality comes with its own challenges and limitations \cite{rahmani_2021_survey}. 
Multimodal HAR methods aim to mitigate the shortcomings of unimodal approaches by fusing information extracted from different sources of data \cite{yadav_2021_survey}. With the breakthrough success of deep learning in the past years, various architectures of deep neural networks have shown impressive performance in multimodal HAR. Nevertheless, they have a common significant drawback, namely, they require vast amounts of labeled data for training deep models. 

\begin{figure}[!t]
\centering
\scalebox{0.7}{
\includegraphics[width=10cm]{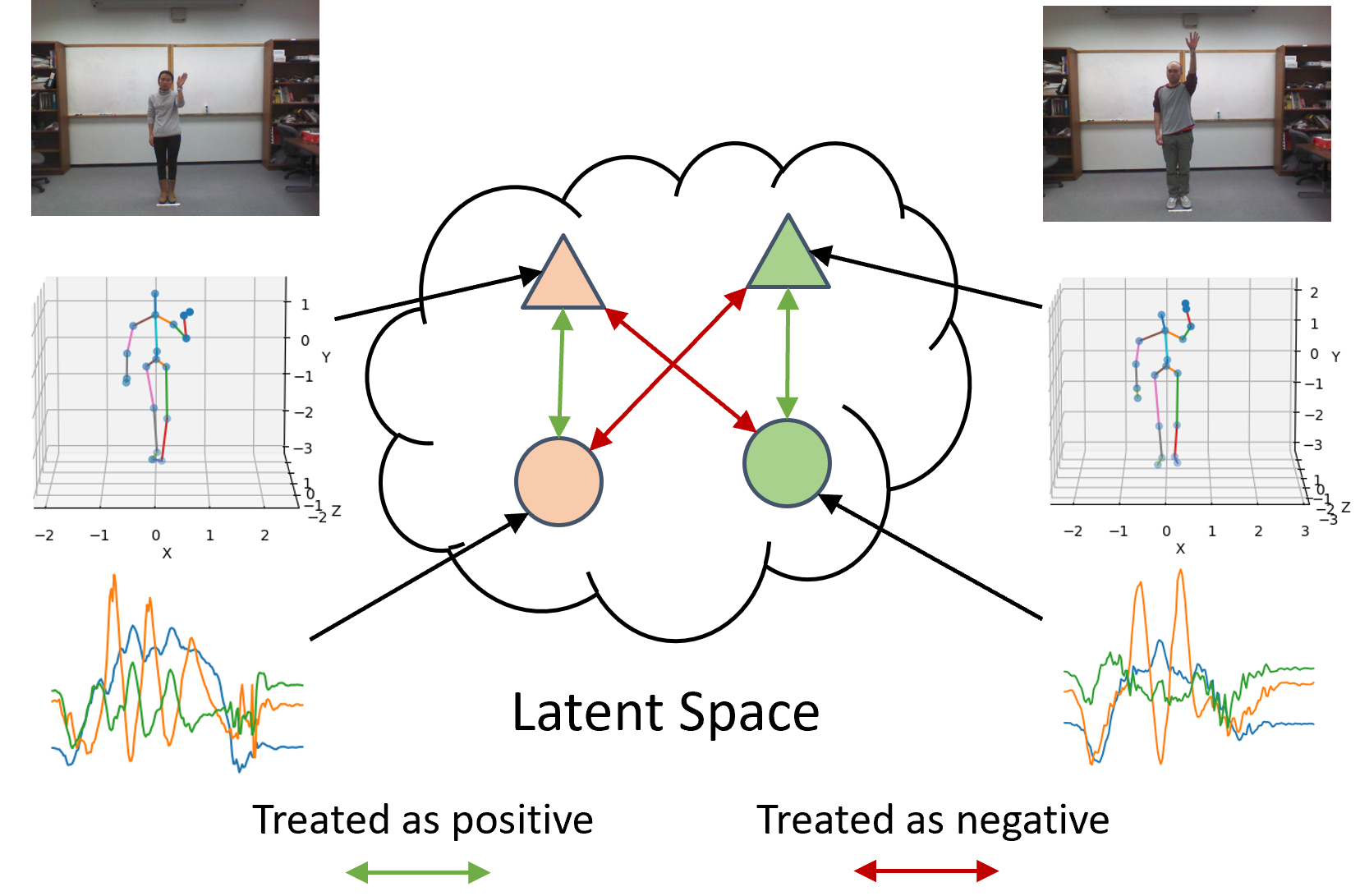}
}
\caption{Illustrated example of a false negative pair in contrastive learning for multimodal HAR. Even though both examples belong to the waving activity class, they are treated as a negative pair, since annotations are not available.}
\label{fig:false_neg}
\vspace{-15pt}
\end{figure}

Given that labeled data is limited and hard to generate, being able to train a HAR model on unlabeled data could have real practical applications. This is the idea behind the self-supervised learning (SSL) paradigm. Specifically, SSL models aim to learn robust feature representations by solving an auxiliary task which can be defined entirely in the unlabeled setting. This process is also known as self-supervised pre-training, and the auxiliary tasks are commonly named pre-text tasks. Then, during the fine-tuning stage, the obtained representations are used to train a shallow classification model for the original (downstream) task using limited amounts of annotated data.



SSL methods have been successfully applied to HAR using individual modalities, but the multimodal setting is still insufficiently explored. Most recent self-supervised models in visual or sensor domains rely on a contrastive learning objective that aims to project the raw inputs into a feature space, such that similar, or positive, sample pairs have close representations, while semantically different, or negative, pairs are spaced apart. In multimodal settings, the Contrastive Multiview Coding (CMC) framework \cite{Tian_2020_cmc} forms positive pairs between the different modalities of each data sample, and negative pairs between different modalities of different samples. Since the data is unannotated, this implies that datapoints from negative pairs might correspond to the same class label in the downstream task \cite{Chuang_2020_debiased}. 
The presence of these false negatives is one of the major drawbacks of self-supervised contrastive learning approaches which rely on negative pairs. We visualize this limitation for inertial and skeleton modalities in Figure \ref{fig:false_neg}. Furthermore, as evidenced in \cite{Zolfaghari_2021_crossclr}, CMC uses only inter-modality negatives, although employing intra-modality negatives as well might have a positive impact on the intra-modal alignment of features. 

In this paper, we aim not only to adapt contrastive learning to multimodal Human Activity Recognition using wearable sensor and skeletal data but also mitigate the limitations of the classical contrastive learning approaches by introducing a Contrastive Multiview Coding with Cross-Modal Knowledge Mining (CMC-CMKM) framework. 
The main contributions of this work are listed as follows: 

\begin{itemize}
    \item We implement the contrastive multiview coding (CMC) algorithm \cite{Tian_2020_cmc} to extract robust feature representations from inertial and skeleton data in the SSL settings. Moreover, we compare the designed models with the supervised and unimodal SSL approaches, namely SimCLR, built for each modality independently.
    \item We address the problem of false negative samples by introducing cross-modal knowledge mining techniques. First, we propose using feature representations learnt by unimodal encoders to mine additional positive pairs for the CMC framework, assuming they might otherwise represent false negatives. Besides, we propose using intra-modality positives and negatives to enhance the intra-modal alignment of features.
    \item Extensive experiments have been carried out on two open-source datasets containing inertial and skeleton modalities, namely UTD-MHAD \cite{utd_mhad} and MMAct \cite{Kong_2019_mmact}. 
\end{itemize}

\section{Related work}
\subsection{Unimodal Human Activity Recognition}
The most widely-used approaches for performing human activity recognition on inertial data often use CNNs \cite{yang_2015_cnn_sensor} or RNNs \cite{Hammerla_2016} or a combination of these types of networks \cite{xu_2019_innohar}. Recent works have also explored more advanced architectures based on attention mechanisms\cite{Zeng_2018, Mahmud_2020}, and deep metric learning, which attempt to learn robust feature embeddings relying on various contrastive loss functions, such as triplet loss \cite{bulat_2021_triplet}, in a supervised manner.

Another widely-used input modality for HAR is skeleton data. A powerful technique for processing skeleton data is co-occurrence feature learning, which transposes the data in different ways as it is passed through a CNN, to capture both temporal and spatial relations between joints \cite{li_2018_cooccurrence}. Recurrent neural networks have also been proposed as an alternative architecture for classifying skeleton sequences \cite{song_2017_stlstm}. More recently, a great deal of attention has been given to graph convolutional networks, which explore the intuition of representing spatial and temporal structure of skeleton data as graphs \cite{yan_2018_stgcn, cheng_2020_shiftgcn, song_2021_efficientgcn}. 


\subsection{Multimodal Human Activity Recognition}

While promising results have been obtained by leveraging individual modalities for HAR, each modality has its own limitations. Then, combining multiple modalities should lead to more robust predictions in practical use cases. One of the main challenges in performing multimodal HAR is combining (or fusing) all of this different information in a coherent way, in order to provide a single prediction.

\begin{figure*}[!t]
\centering
\scalebox{0.9}{
\subfloat[CMC pre-text task]{\includegraphics[width=0.578\textwidth, keepaspectratio]{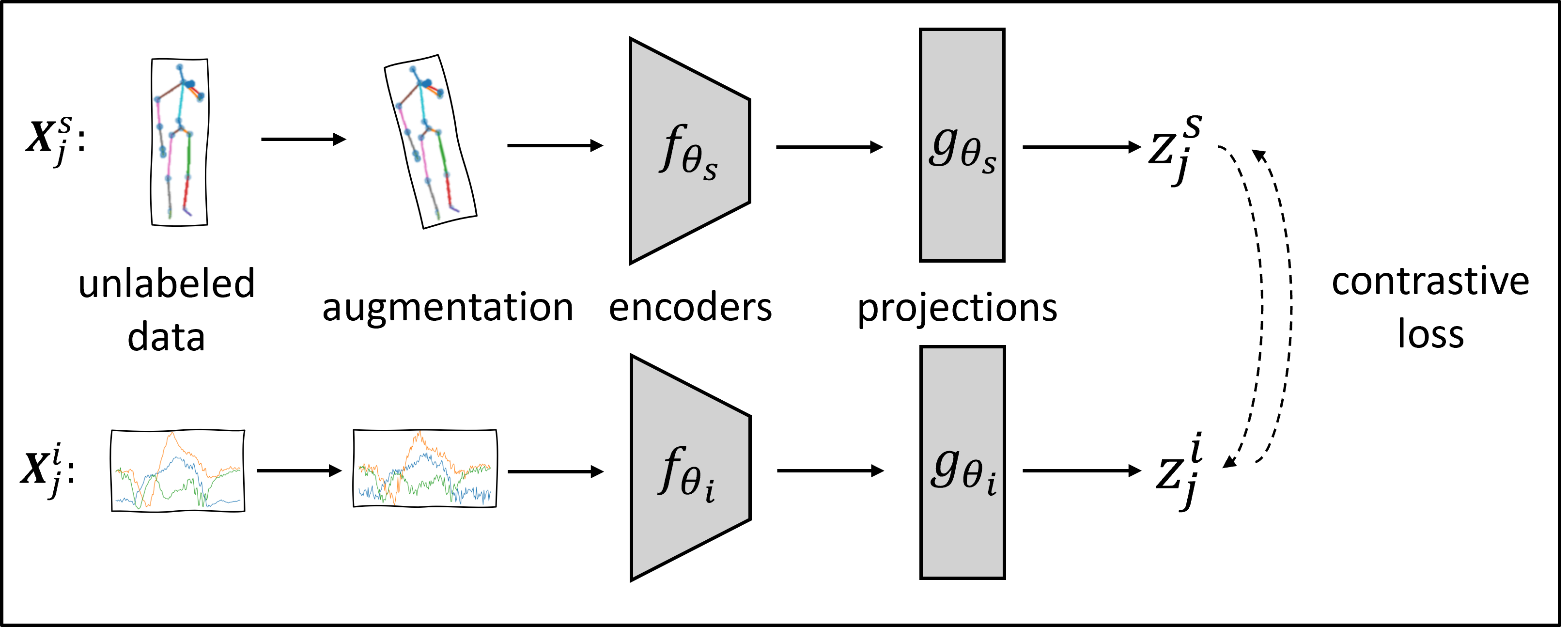}\label{fig:cmc}}
\hspace{\fill}
\subfloat[Fine-tuning routine]{\includegraphics[width=0.398\textwidth, keepaspectratio]{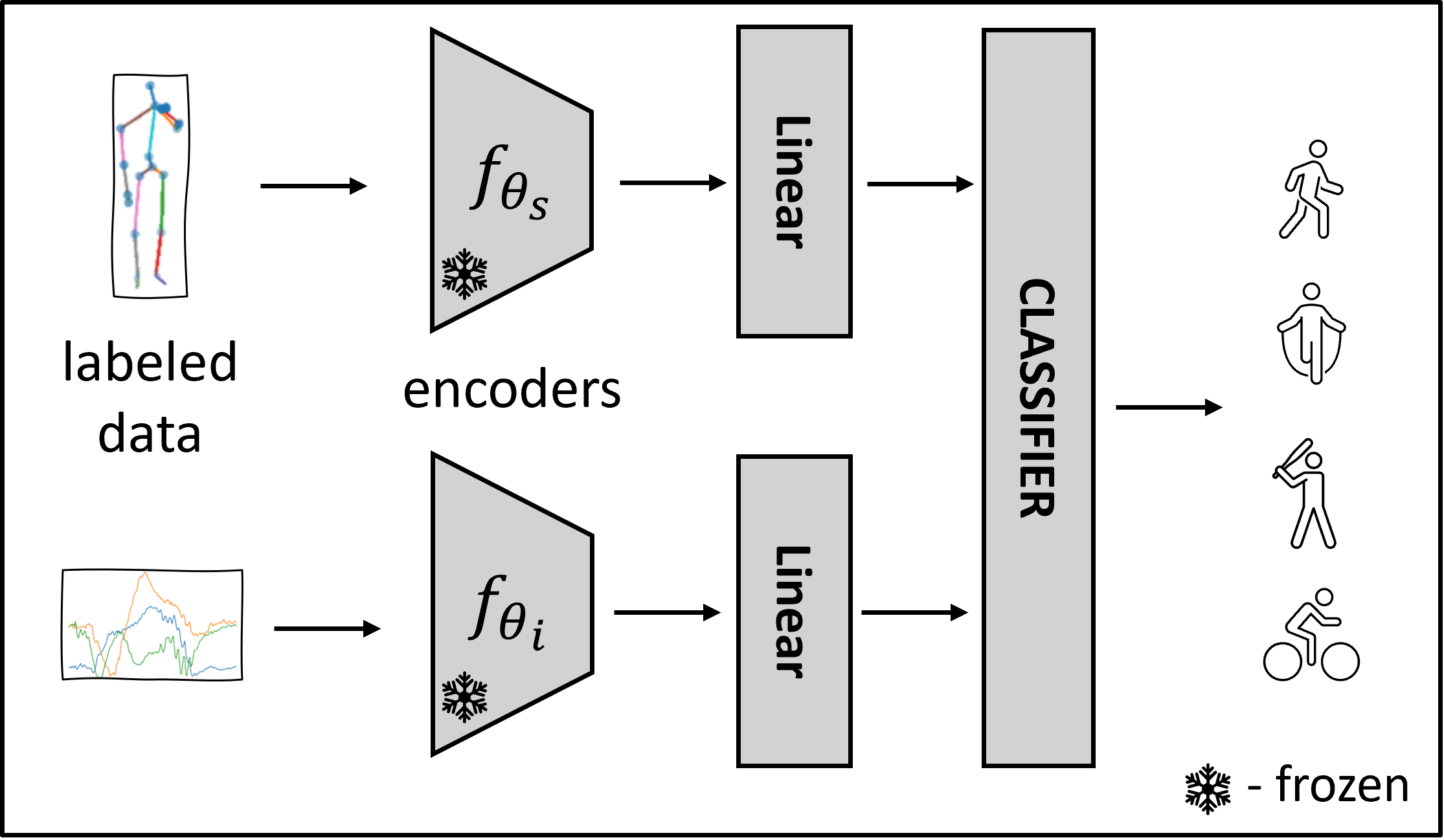}\label{fig:finetuning}}}
\caption[]{Multimodal Contrastive Learning stages: pre-training (a) - inertial and skeleton data examples are passed through the modality-specific encoders and projection heads to generate representations used in contrastive loss; fine-tuning (b) - the pre-trained frozen encoders produce features for labeled data, these features are then passed through a mapping linear layer and classifier to get activity labels.}
\label{fig:vanilla_lstm}
\vspace{-10pt}
\end{figure*}

To account for the significant difference between input modalities, many multimodal HAR works apply input, feature or decision fusion in various architectures comprising of multiple backbone networks suitable for individual modalities \cite{khaire_2018_combiningcnn, memmesheimer_2020_gimme_signals, das_2021_mmhar_ensemnet}. More sophisticated recent works propose end-to-end architectures designed specifically for multimodal HAR. For example, Wang et al. propose a multi-view generative framework where GANs are used to generate the feature encodings of one view, given another \cite{wang_2019_generative_har}. This allows their framework to be used for inference even when one of the original modalities is unavailable. In \cite{islam_2020_hamlet}, multiple multi-head attention mechanisms are used to encode and fuse features from different modalities. 
Liu et al. \cite{liu_2021_sakdn} proposed a framework which preserves the semantics of the original data by distilling knowledge from a teacher network which is trained on inertial data, to a student network which only uses RGB data. 

\subsection{Contrastive learning for Human Activity Recognition}


Recent contrastive SSL methods rely on maximising the latent similarity of augmented views originating from the same data sample \cite{he_2020_moco, Chen_2020_simclr}. The main issue associated with contrastive learning frameworks is their reliance on negative pairs. Besides the fact that contrastive pre-text tasks which use negative pairs normally require large batch sizes, there is also the issue of false negative pairs which may harm learning. While some works proposed using positive pairs only by introducing additional constraints to avoid trivial solutions \cite{grill_2020_byol, chen_2021_simsiam}, others suggested different approaches to mitigate the impact of false negatives \cite{Chuang_2020_debiased, Li_2021_crossclr, Zolfaghari_2021_crossclr}.


In the field of HAR, contrastive SSL has mainly been applied on individual data sources, such as sensors \cite{haresamudram_2021_cpc_har, Khaertdinov_2021_csshar}, skeleton data \cite{Li_2021_crossclr} or visual data \cite{han_2020_cotraining}. Despite the large number of supervised techniques which have achieved good performance on multimodal HAR, very few works, to the best of our knowledge, have addressed this problem using self-supervised learning. Akbari et al. \cite{akbari_2021_vatt} introduced the VATT framework that uses modality-specific and modality-agnostic Transformer encoders for multimodal self-supervised learning using video, audio and text modalities.

In this paper, we adapt the Contrastive Multiview Coding framework \cite{Tian_2020_cmc}, previously used in Computer Vision applications, to the problem of multimodal HAR using inertial and skeleton data. Moreover, inspired by ideas suggested in \cite{Li_2021_crossclr} and \cite{Zolfaghari_2021_crossclr}, we introduce a novel cross-modal knowledge mining technique that can be easily plugged into the the CMC framework (CMC-CMKM). It aims to mitigate the impact of false negative pairs by using knowledge from each modality to guide the training process.

\section{Methodology}

\subsection{Problem Definition}
Multimodal Human Activity Recognition can be formulated as a classification problem where, given a set of inputs $\{\boldsymbol{X}^m \, \vert \, m \in M \}$ from a set of modalities $M$, the objective is to predict the label $y \in Y$ associated with these inputs. The remainder of this paper will focus on two input modalities, namely inertial (or sensor) data and skeleton data.

Inertial signals are generally obtained from wearable devices such as accelerometers, magnetometers or gyroscopes, and have the shape of multivariate time series. At any timestamp $t$, the input signal $\boldsymbol{x_t} = [x_t^1, x_t^2, \ldots, x_t^S] \in \mathbb{R}^{S}$ consists of $S$ values obtained from the S available sensor channels. In matrix form, an inertial data sample recorded over T timestamps is denoted as $\boldsymbol{X}^i = [\boldsymbol{x_1}, \boldsymbol{x_2}, \ldots, \boldsymbol{x_T}] \in \mathbb{R}^{T \times S}$.

Skeleton data is generally provided as a set of 2D or 3D coordinates tracked over time for a number of keypoints located on the human body. 
For any skeleton sequence, we denote $T$ as the number of frames in the sequence, $J$ as the number of joints and $C$ as the number of data channels (2 or 3). Then, a skeleton sequence $\boldsymbol{X}^s \in \mathbb{R}^{T \times J \times C}$ consists of $T$ frames where the skeleton data for each frame is described by $\boldsymbol{P}_t = [\boldsymbol{p}_t^1, \boldsymbol{p}_t^2, \ldots, \boldsymbol{p}_t^J]$ and $\boldsymbol{p}_t^j \in \mathbb{R}^C$ is the position of joint $j$ at frame $t$.

\subsection{Contrastive Learning for Multimodal HAR}
The first contribution of this paper is the adaptation of the widely-used Contrastive Multiview Coding framework to the problem of multimodal HAR. It is a contrastive self-supervised learning method which can be used when two or more representations per example are available for each sample \cite{Tian_2020_cmc}. Specifically, the proposed adaptation of CMC contrasts between the feature embeddings obtained from the inertial and skeleton modalities.

\begin{figure*}[!t]
\centering
\scalebox{1}{
\includegraphics[width=15cm]{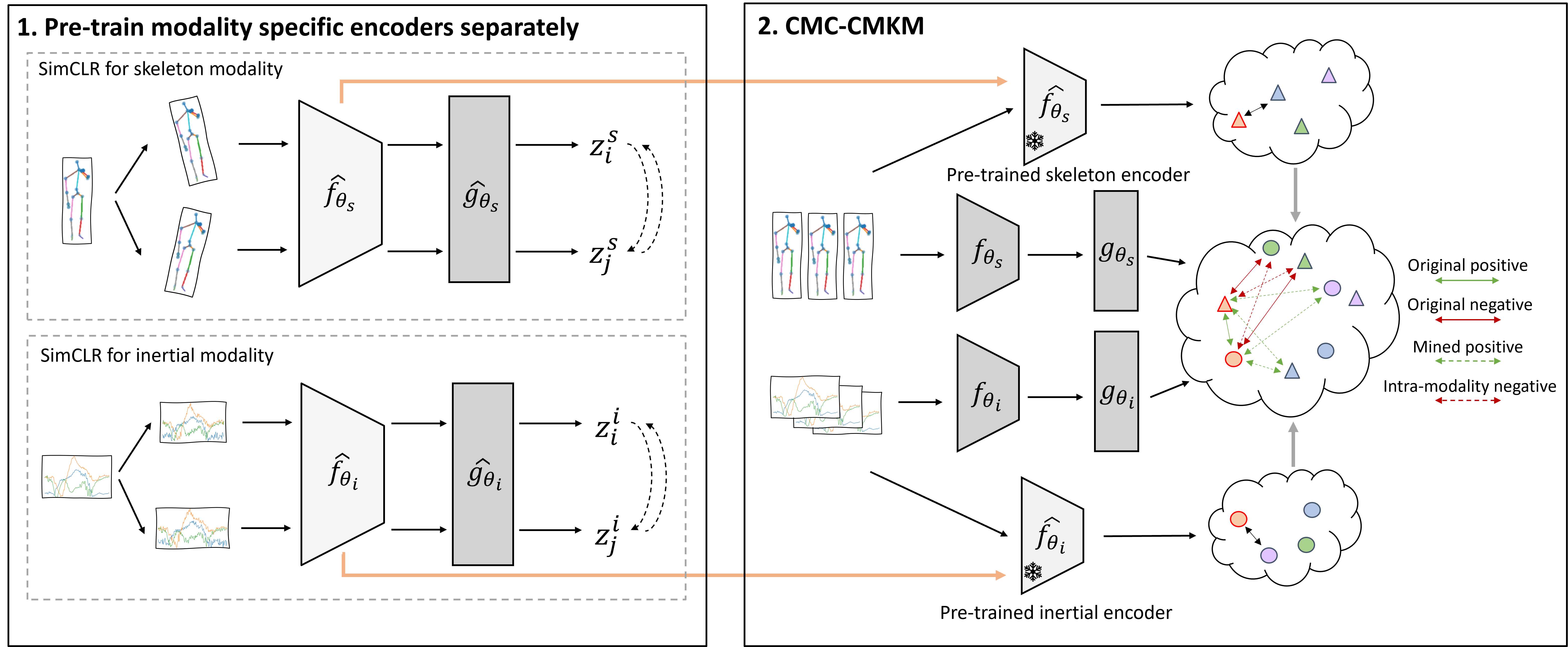}
}
\caption{CMC with Cross-modal knowledge mining (CMKM). First, additional encoders are pre-trained separately using SimCLR. Second, the additional encoders are used to guide additional positive mining within the CMC framework. Besides, CMC-CMKM uses intra-modalilty negatives. Triangles and circles indicate skeleton and inertial features, respectively, while each color corresponds to one instance. For the orange instance, a number of positive and negative relationships are shown.}
\label{fig:cmkm}
\vspace{-15pt}
\end{figure*}

Formally, for each sample $\{\boldsymbol{X}_j^{i},\boldsymbol{X}_j^{s}\}$ in a training batch of size $N$ (where $i$ and $s$ refer to the inertial and skeleton modalities, respectively), the input data for each modality is augmented with a random modality-specific augmentation, generating $\tilde{\boldsymbol{X}}^i_j=t_j^i(\boldsymbol{X}_j^{i})$ and $\tilde{\boldsymbol{X}}^s_j=t_j^s(\boldsymbol{X}_j^{s})$. For CMC, the purpose of data augmentation is simply to improve learning by extending the size of the dataset. Then, two modality-specific encoders $f_{\theta_i}, f_{\theta_s}$ and projection heads $g_{\theta_i}, g_{\theta_s}$ are used to generate projections $\boldsymbol{z}_j^{i}=g_{\theta_i}(f_{\theta_i}(\tilde{\boldsymbol{X}}^i_j))$ and $\boldsymbol{z}_j^{s}=g_{\theta_s}(f_{\theta_s}(\tilde{\boldsymbol{X}}^s_j))$. These representations are then treated as a positive pair. This process is illustrated in Figure \ref{fig:cmc}. The negative pairs are formed by all inter-modal combinations of projections which do not originate from the same input instance. Thus, the loss obtained by treating $\boldsymbol{X}_j^{i}$ as an anchor and enumerating over the representations of the other samples $\boldsymbol{X}_k^{s}$ is:


\begin{equation}
    l_j^{i \rightarrow s} = - log \frac{\delta({z}_j^{i},{z}_j^{s})}{\sum_{k=1}^N \delta({z}_j^{i},{z}_k^{s})},
    \label{eq:info-nce}
\end{equation}

where $\delta({z}_j^{i},{z}_j^{s}) = exp(s({z}_j^{i},{z}_j^{s}))/\tau$ and $s(\cdot)$ is the cosine similarity function.

The total loss accumulated over the training batch is calculated as follows:

\begin{equation}
    \mathcal{L} = \sum_{j=1}^N (l_j^{i \rightarrow s} + l_j^{s \rightarrow i})
    \label{eq:cmc_total_loss}
\end{equation}

The inertial and skeleton encoders pre-trained within the CMC framework are then frozen and used in the fine-tuning stage as shown in Figure \ref{fig:finetuning}. Specifically, we map inertial and skeleton features to the same size using a single fusion linear layer, including batch normalization and ReLU, concatenate the outputs and pass the resulting feature vector through the classification model.

To provide a comparison with CMC, we have also implemented the SimCLR \cite{Chen_2020_simclr} framework for both modalities independently. Specifically, the pre-text task is performed separately for the inertial and skeleton encoders. In this framework, two random sets of augmentations are applied to each input instance to create positive pairs. 

\subsection{Cross-Modal Knowledge Mining}
Contrastive multiview coding is a powerful framework for performing multimodal SSL. However, the training procedure and the formulation of the loss function still rely on a set of underlying assumptions which might have a negative impact on the learned representations. First, CMC relies heavily on negative pairs in its contrastive objective, which makes it more sensitive to batch size. Furthermore, as ground-truth labels are unavailable during the pre-training process, there is no way to prevent the negative pairs from occasionally consisting of false negatives. Finally, as CMC only contrasts between representations from different modalities, the encoders are not explicitly trained to preserve intra-modal similarities.

First, inspired by \cite{Li_2021_crossclr}, we upgrade the CMC framework with a novel method for cross-modal and intra-modal positive mining. We take advantage of the cross-modal setting and use knowledge from one modality to guide the training process for the other. Intuitively, if two samples are very similar in one modality, there is a chance that the samples might come from the same underlying action class. Applying this intuition to our framework, we use similarities between representations learnt by encoders pre-trained separately using SimCLR to mine additional positives. Specifically, for each modality, we use a pre-trained encoder to compute a similarity matrix containing the pair-wise similarity values for each pair of sample encodings in a batch. For each instance, we select the top-\textit{K} most similar samples, for a fixed \textit{K} value, and we include these mined samples in the positive sets of the instance, in both modalities. 

More formally, given a training batch of size $N$, and intra-modality similarity matrices $\boldsymbol{S}^i$ and $\boldsymbol{S}^s$ for the inertial and skeleton modalities, respectively, we define the positive and negative sets as follows:

For the inertial modality:
\[ P_j^i = \{z_j^s\} \cup \{z_l^s \in TopK(\textbf{S}^s_j) \} \cup \{z_l^i \in TopK(\textbf{S}^i_j) \} \]
\[ N_j^i = \{ z_k^s \, \vert \, 0 \leq k \leq N \} \setminus P_j^i  \]

For the skeleton modality:
\[ P_j^s = \{z_j^i\} \cup \{z_l^i \in TopK(\textbf{S}^i_j) \} \cup \{z_l^s \in TopK(\textbf{S}^s_j) \} \]
\[ N_j^s = \{ z_k^i \, \vert \, 0 \leq k \leq N \} \setminus P_j^s  \]

We note that both positive sets contain embeddings originating from both modalities. The respective loss term for each sample in the inertial modality becomes:

\begin{equation}
    l_j^{i \rightarrow s} = - log \frac{\sum_{z_k^m \in P_j^i} \delta(z_j^i, z_k^m)}{\sum_{z_k^m \in P_j^i \cup N_j^i} \delta(z_j^i, z_k^m)}
\end{equation}


The loss term for the skeleton modality is defined according to the same logic.

Additionally, we exploit intra-modality negatives to encourage the model to better align features inside each modality. This is done by adding the similarities of intra-modality negatives to the denominator of the loss term. Finally, the proposed loss function for each sample can be formally written as follows:

\begin{equation}
    l_j^{i \rightarrow s} = - log \frac{\sum_{z_k^m \in P_j^i} \delta(z_j^i, z_k^m)}{\sum_{z_k^m \in P_j^i \cup N_j^i} \delta(z_j^i, z_k^m) + \sum_{z_k^i \in N_j^s} \delta(z_j^i, z_k^i)}
\label{eq:final_loss}
\end{equation}


We also note that the addition of intra-modality negatives is consistent with positive mining, as the mined samples are also implicitly excluded from the intra-modality negative sets. An illustrative example of the proposed improvements is shown in Figure \ref{fig:cmkm}. The whole pre-training routine is summarized in Algorithm \ref{alg:cmkm}. 

\subsection{Backbone Models}
The proposed framework requires two encoders to extract features from inertial and skeleton signals. For inertial data, the encoder $f_{\theta_i}$ is the transformer-like encoder described in the CSSHAR framework \cite{Khaertdinov_2021_csshar}. The input data is passed through a one-dimensional CNN with batch normalization and a ReLU non-linearity, then through a positional encoding layer and a transformer encoder consisting of multiple self-attention blocks, as described in the original Transformer architecture \cite{vaswani_2017_transformer}. 

For the skeleton modality, we picked the lightweight convolutional co-occurrence feature learning network \cite{li_2018_cooccurrence}. It uses a two-stream input (of positions and motions) and comprises of a series of convolutional blocks, with ReLU non-linearities and max-pooling applied to certain layers. A key element of this architecture is a transpose block which is inserted into the network between two intermediate layers, and which rearranges the data such that the joints become the input channels of subsequent convolutions. This allows the network to learn features in a hierarchical manner, from point-level features describing each joint, to co-occurrence features which capture the relationship between the different joints in a sequence.

\begin{algorithm}[!t]
\SetAlgoLined
\KwData{unlabelled dataset $\{\boldsymbol{X}^i_k, \boldsymbol{X}^s_k\}^N_{k=1}$, where $N$ is the number of training samples}

\KwIn {inertial encoders $\hat{f_{\theta_i}}, f_{\theta_i}$ and projections heads $\hat{g_{\theta_i}}, g_{\theta_i}$, skeleton encoders $\hat{f_{\theta_s}}, f_{\theta_s}$ and projections heads $\hat{g_{\theta_s}}, g_{\theta_s}$}

\BlankLine
\# \textit{stage 1: unimodal pre-training}
 
    pre-train encoder $\hat{f_{\theta_i}}$ and projection head $\hat{g_{\theta_i}}$ using SimCLR, then discard $\hat{g_{\theta_i}}$ and freeze $\hat{f_{\theta_i}}$\;
 
    pre-train encoder $\hat{f_{\theta_s}}$ and projection head $\hat{g_{\theta_s}}$ using SimCLR, then discard $\hat{g_{\theta_s}}$ and freeze $\hat{f_{\theta_s}}$\;
 
\BlankLine
\# \textit{stage 2: main multimodal pre-training}

\For {each training batch $\{\boldsymbol{X}^i_k, \boldsymbol{X}^s_k\}^n_{k=1}$}{
    
    obtain augmented samples $\{\tilde{\boldsymbol{X}}^i_k, \tilde{\boldsymbol{X}}^s_k\}^n_{k=1}$\;
    
    compute projections $\{z^i_k\}^n_{k=1}$, $\{z^s_k\}^n_{k=1}$\;

    \BlankLine
    \# \textit{positive mining}
    
    
    for $k \in \{1,...,n\}, l \in \{1,...,n\}$, compute $\textbf{S}_{k,l}^i = s(\hat{f_{\theta_i}}(\tilde{\boldsymbol{X}}^i_k), \hat{f_{\theta_i}}(\tilde{\boldsymbol{X}}^i_l))$, $\textbf{S}_{k,l}^s = s(\hat{f_{\theta_s}}(\tilde{\boldsymbol{X}}^s_k), \hat{f_{\theta_s}}(\tilde{\boldsymbol{X}}^s_l))$\;
    
    
    define sets $P_j^i, N_j^i, P_j^s, N_j^s$\;
    
    \BlankLine
    \# \textit{contrastive loss}
    
    \For {$k \in \{1, ..., n\}$}{
        compute $l_k^{i \rightarrow s}, l_k^{s \rightarrow i}$ according to Equation \ref{eq:final_loss}\;
    }
    
    compute total loss $\mathcal{L} = \sum_{k=1}^N (l_k^{i \rightarrow s} + l_k^{s \rightarrow i})$\;
    
    update $f_{\theta_i}, g_{\theta_i}, f_{\theta_s}, g_{\theta_s}$ to minimize $\mathcal{L}$\;
}
        
 \caption{Model pre-training using cross-modal knowledge mining}
 \label{alg:cmkm}
\end{algorithm}
\section{Implementation details}
\subsection{Datasets}
In this paper, two open-source multimodal datasets were used to evaluate the performance of the proposed approaches, namely UTD-MHAD \cite{utd_mhad} and MMAct \cite{Kong_2019_mmact}. Skeleton and inertial modalities were extracted and used from both datasets.

\noindent\textbf{UTD-MHAD.} The dataset contains data collected by 10 subjects performing 27 activities, 4 trials for each. The three-dimensional joint coordinates were recorded with a Kinect camera, while the inertial data was collected using one wearable device with accelerometer and gyroscope. We follow the original evaluation protocol, using odd-numbered subjects for training and even-numbered subjects for testing and reporting test accuracy.

\noindent\textbf{MMAct.} The dataset consists of 36 activities performed by 20 subjects in different scenes. For skeleton data, we employ the 2D keypoints present in the challenge version of the dataset\footnote{challenge dataset: \url{https://mmact19.github.io/challenge/} }. The sensor data comes from a smartwatch (accelerometer) and a smartphone (accelerometer, gyroscope, orientation) located in the subject's pocket. We follow the cross-subject and cross-scene evaluation protocols. Specifically, for the former we use the samples from the first 16 subjects for training and the others for testing, while for the latter we reserve all samples collected in the occlusion scene for testing, and train on all subjects and all other scenes. As per the authors' recommendation, we report the F1 score obtained on the test set.

\subsection{Hyperparameters}
In this subsection, we describe the hyperparameters used to pre-train and fine-tune the proposed models as well as specific details regarding the architectures of the modality-specific encoders. To optimize the parameters of the models, we use the Adam optimizer with a learning rate of $0.001$ which is reduced twice when learning stagnates for more than 20 epochs.

\noindent\textbf{Inertial encoder.}
The inertial encoder implemented in this paper is adapted from CSSHAR \cite{Khaertdinov_2021_csshar}. Specifically, first, input data is passed through $3$ one-dimensional CNN blocks consisting of $[32, 64, 128]$ feature maps with kernel size $5$. The obtained feature maps are then used as an input for $2$ self-attention blocks with $2$ heads each.

\noindent\textbf{Skeleton encoder.} The skeleton encoder adopted for the experiments follows a hierarchical co-occurrence learning architecture \cite{li_2018_cooccurrence}. We implement the model as described in the original paper, only replacing dropout layers with batch normalization layers.  

\noindent\textbf{Initial pre-processing.} We re-sample all input sequences to 50 timesteps, for both inertial and skeleton data. Additionally, we normalise joint positions in all skeleton sequences based on the first frame of each sequence, following a standard normalisation procedure \cite{khaire_2018_combiningcnn}.

\noindent\textbf{Pre-training.} Prior to pre-training the models, we apply a set of random augmentations for inertial and skeleton modalities. The inertial augmentations, as proposed in \cite{Khaertdinov_2021_csshar}, applied to each input instance are sampled randomly (with 75\% probability) from the set of augmentations \{jittering, scaling, rotation\} for UTD and \{jittering, scaling, permutation, channel shuffle\} for MMAct. For the skeleton modality, we use \{jittering, random resized crops, scaling, rotation, shearing\} on both datasets. Jittering is always applied, while the other augmentations are applied with a 75\% probability. We pre-train with SimCLR for 300 epochs, and with CMC-CMKM for 100 epochs. For the SimCLR experiments on inertial data, we use batch sizes of 128 and 64 and temperature values of 0.05 and 0.2 for UTD-MHAD and MMAct, respectively. For skeleton SimCLR, we use batches of 32 and 128 samples, and temperatures of 0.5 and 0.2. For the multimodal experiments using CMC-CMKM, we used a batch size of 64 for UTD-MHAD and 128 for MMAct, and temperature values of 0.1 for both datasets.

\noindent\textbf{Fine-tuning.} For the fine-tuning routine, which remains the same for all multimodal approaches (Figure \ref{fig:finetuning}), we implement modality-specific fusion layers ($1$ layer per modality), including batch normalization and ReLU, that map inertial and skeleton embeddings to the same size of 256. The obtained features are then concatenated and passed through a simple linear classifier with softmax activation. We train the modality specific fusion layers and linear classifier for 100 epochs using the labels of the downstream task.
\section{Evaluations}
\begin{table}[!t]
\centering
\begin{tabular}{cccc}
                       & UTD-MHAD   & \multicolumn{2}{c}{MMAct (F1-score)} \\
top-$K$                   & (Accuracy) & x-subject         & x-scene          \\ \hline
\multicolumn{1}{c}{0} & 94.88      & 83.36             & 79.06            \\
1                      & \textbf{97.67}      & \textbf{84.51}    & \textbf{82.91}   \\
2                      & 96.05      & 81.92             & 81.73            \\
3                      & 96.05      & 82.41             & 82.77            \\
4                      & 94.65      & 82.64             & 81.4             \\
5                      & 94.88      & 82.96             & 82.84           
\end{tabular}
\caption{Ablation for different $k$ in cross-modal positive mining. The row with $k=0$ refers to the model using intra-modality negatives only.}
\label{tab:topk}
\end{table}

Our code has been made publicly available on GitHub\footnote{https://github.com/razzu/cmc-cmkm}. All experiments have been run on a single Nvidia Quadro RTX 5000 card. One epoch of CMC pre-training on MMAct takes approximately 8-9 seconds, while one epoch of CMC-CMKM pre-training takes 14-15 seconds. It is also worth mentioning that CMC-CMKM pre-training requires models pre-trained in the unimodal settings, with unimodal pre-training taking approximately 15-17 seconds per epoch. Finally, fine-tuning for both CMC and CMC-CMKM takes approximately 4-5 seconds per epoch.

\subsection{Learning Feature Representations}
In order to evaluate the representations learnt by the proposed multimodal approaches, we perform linear evaluation on top of the fused features extracted by the pre-trained modality-specific encoders. Specifically, the whole annotated datasets are used to fine-tune the fusion layer and linear classifier as shown in Figure \ref{fig:finetuning}.

\noindent\textbf{Number of mined positives.} First, we explore how the number of mined positives $k$ affects the performance of models in the proposed cross-modal knowledge mining protocol. The results of this experiment are presented in Table \ref{tab:topk}. In Table \ref{tab:topk}, when $K$ is set to 0, only the intra-modality negatives are used. As can be seen from the table, the proposed method shows optimal performance in all three scenarios, when $K$ is equal to 1, meaning that one extra positive is mined from each modality.


\noindent\textbf{Comparison to baselines.} In Table \ref{tab:res_big}, we also compare the performance of the proposed CMC-CMKM approach to other SSL models, pre-trained in the unimodal setting with SimCLR and in the multimodal setting with standard CMC. Additionally, we include the performance of identical encoders trained in a supervised end-to-end manner. 
\begin{table}[!t]
\centering
\scalebox{1}{
\begin{tabular}{ccccc}
           &            & UTD-MHAD       & \multicolumn{2}{c}{MMAct (F1-score)} \\
    Modality   & Approach   & (Accuracy)     & x-subject         & x-scene          \\ \hline
    Inertial   & SimCLR     & 72.09          & 52.89             & 59.23            \\
    Inertial   & Supervised & 76.74          & 61.22             & 78.86            \\ \hline
    Skeleton   & SimCLR     & 95.11          & 75.82             & 67.80            \\
    Skeleton   & Supervised & 94.65          & 82.50             & 70.58            \\ \hline
    Multimodal & CMC        & 96.04          & 82.05             & 79.97            \\
    Multimodal & CMC-CMKM       & \textbf{97.67} & \textbf{84.51}    & \underline{82.91}      \\
    Multimodal & Supervised & \underline{97.21}    & \underline {84.05}       & \textbf{87.36}  
\end{tabular}}
\caption{Linear evaluation results: the highest results are highlighted in bold, the second highest are underlined.}
\label{tab:res_big}
\vspace{-10pt}
\end{table}

According to the obtained performance scores, it is clear that multimodal methods, both SSL and supervised, are much more powerful than the unimodal ones. Furthermore, SSL approaches on multimodal data outperform all the unimodal models trained in the supervised settings. Besides this, the introduced CMC-CMKM approach outperforms CMC by about 2\% in all three scenarios. Moreover, it shows performance comparable to the multimodal supervised model on UTD-MHAD and MMAct (cross-subject) and outperforms it by 0.5\%. For the cross-scene scenario on MMAct, the CMC-CMKM is the closest one to the supervised model.

\begin{figure*}[!h]
    \centering
    \subfloat[UTD-MHAD]{\includegraphics[width=0.3\textwidth, keepaspectratio]{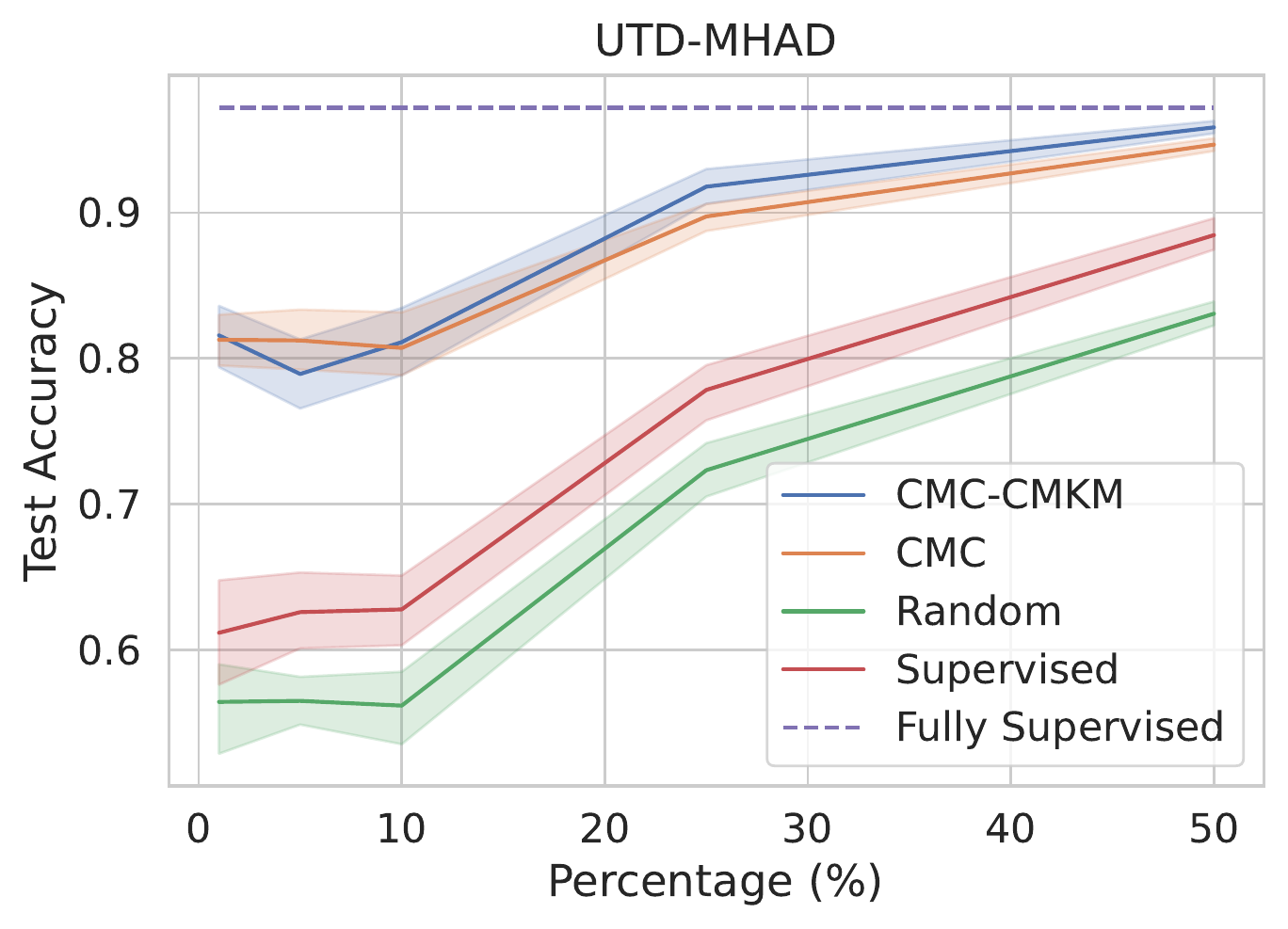}}
    \subfloat[MMAct (x-subject)]{\includegraphics[width=0.3\textwidth, keepaspectratio]{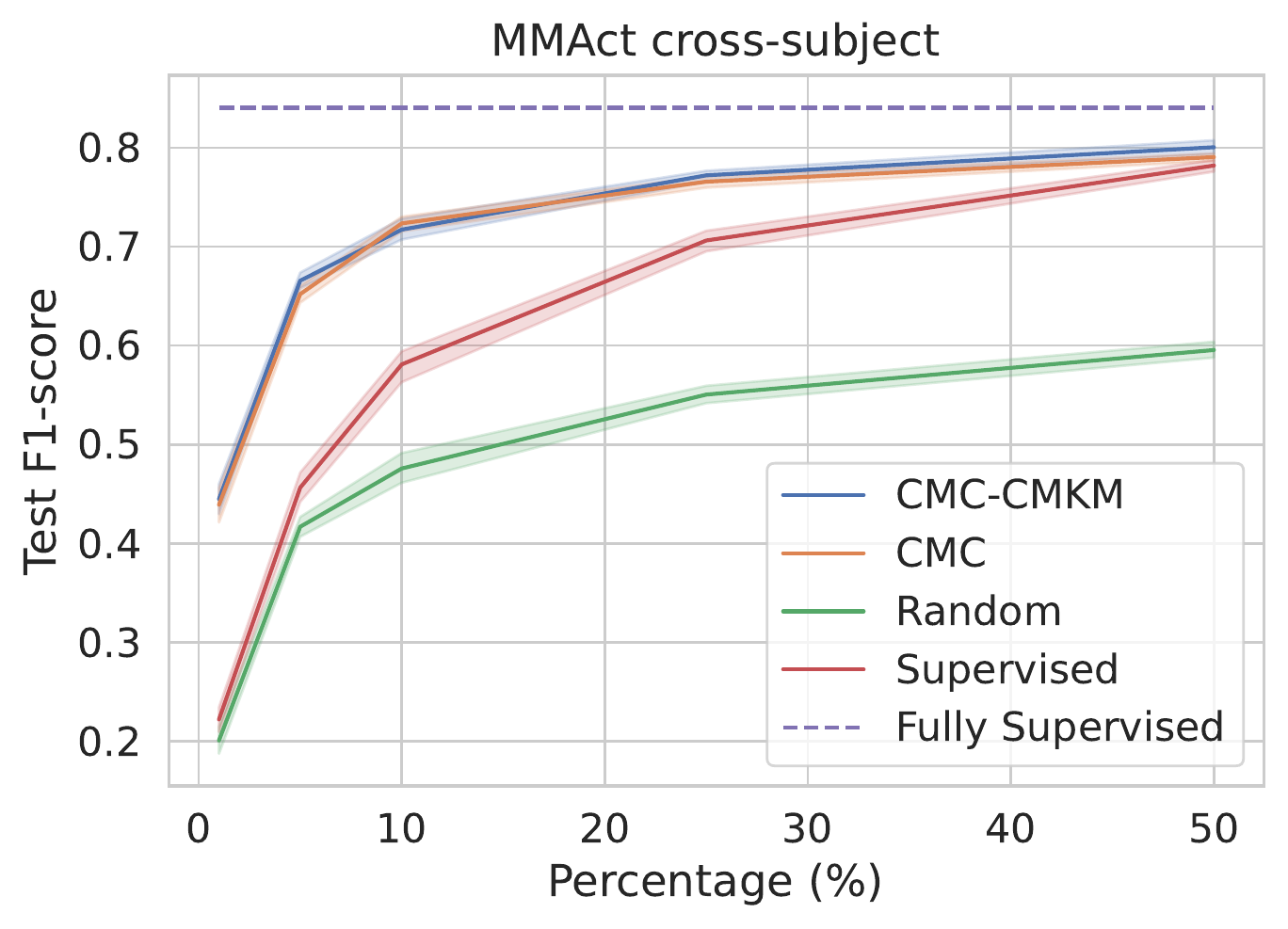}}
    \subfloat[MMAct (x-scene)]{\includegraphics[width=0.3\textwidth, keepaspectratio]{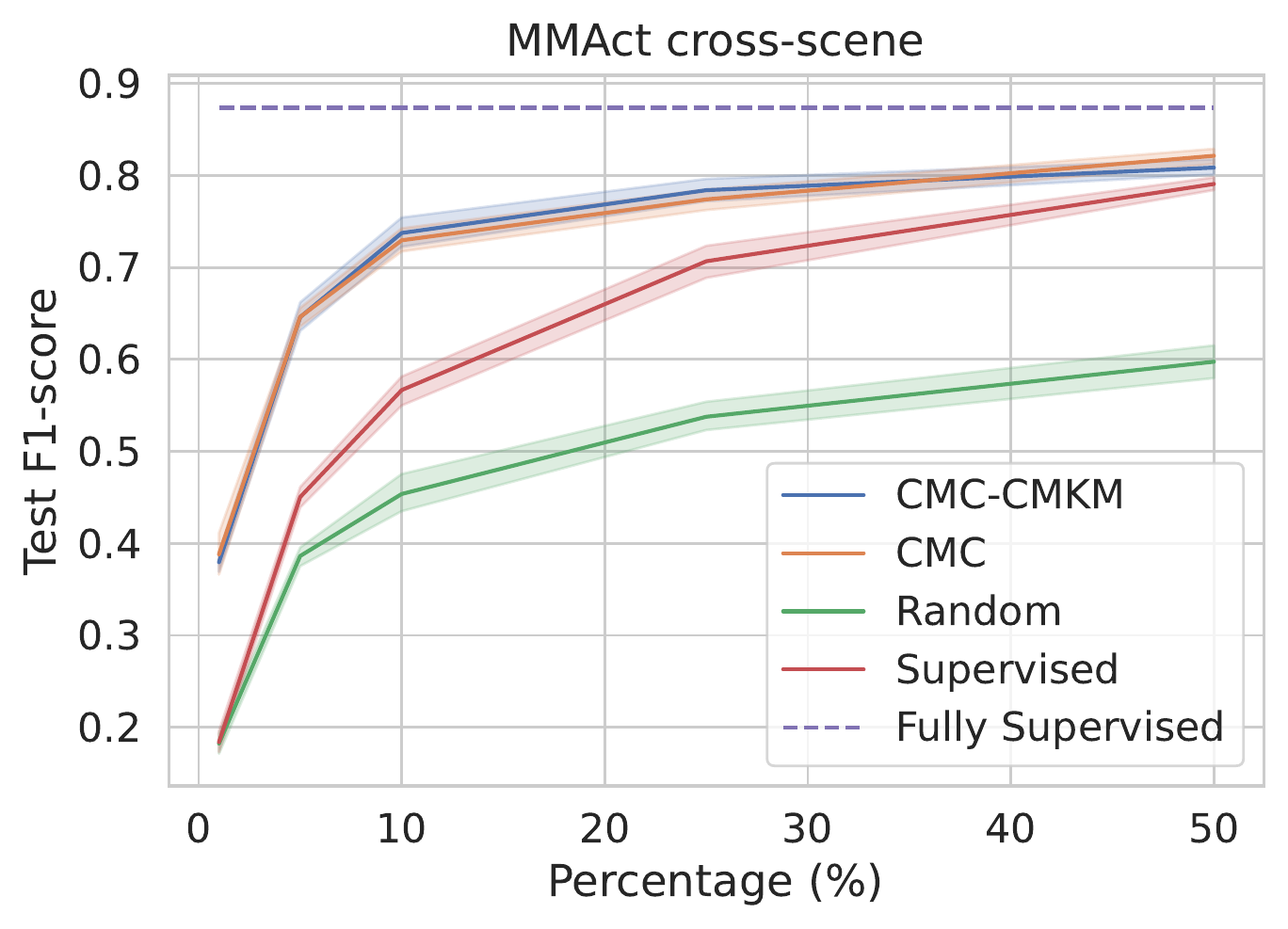}}
    
    \caption[]{Average values of performance metrics with 95\% confidence intervals for the semi-supervised learning scenario.}
    \label{fig:semi_sup}
    \vspace{-10pt}
\end{figure*}

\begin{table}[!t]
\centering
\begin{tabular}{ccccc}
{Pre-text} & \multicolumn{2}{c}{x-subject}   & \multicolumn{2}{c}{x-scene}     \\
                          & inertial       & skeleton       & inertial       & skeleton       \\ \hline
Unimodal SimCLR           & 51.77          & 66.37          & 52.98          & 66.81          \\
CMC                       & 55.26          & 73.98          & 57.33          & 74.26          \\
CMC-CMKM                  & \textbf{56.66} & \textbf{75.77} & \textbf{57.44} & \textbf{75.32}
\end{tabular}
\caption{Activity retrieval accuracy scores on MMAct.}
\label{tab:retrieval}
\vspace{-13pt}
\end{table}

\noindent\textbf{Activity retrieval.} We employ an activity retrieval scenario for modality-specific encoders pre-trained in unimodal and multimodal manner on MMAct. Namely, given an input skeleton or inertial signal stream from the test set, we aim to find the most similar example of the same modality in the training set using learnt feature representations. In this scenario, instead of the default fine-tuning routine, we use kNN ($k = 1$) to predict activities in the test set using inertial and skeleton encoders pre-trained with the contrastive SSL approaches. In other words, we match each feature embedding from the test set with the closest one from the training set using cosine similarity. The accuracy scores for this scenario are shown in Table \ref{tab:retrieval}. According to the obtained results, the encoders pre-trained in multimodal settings significantly outperform models pre-trained in the unimodal manner. Moreover, the proposed CMC-CMKM methods demonstrates better performance than the original CMC indicating improved intra-modal feature alignment for both modalities. 

\begin{figure}[!htbp]
\centering
\scalebox{0.8}{
\includegraphics[width=10cm]{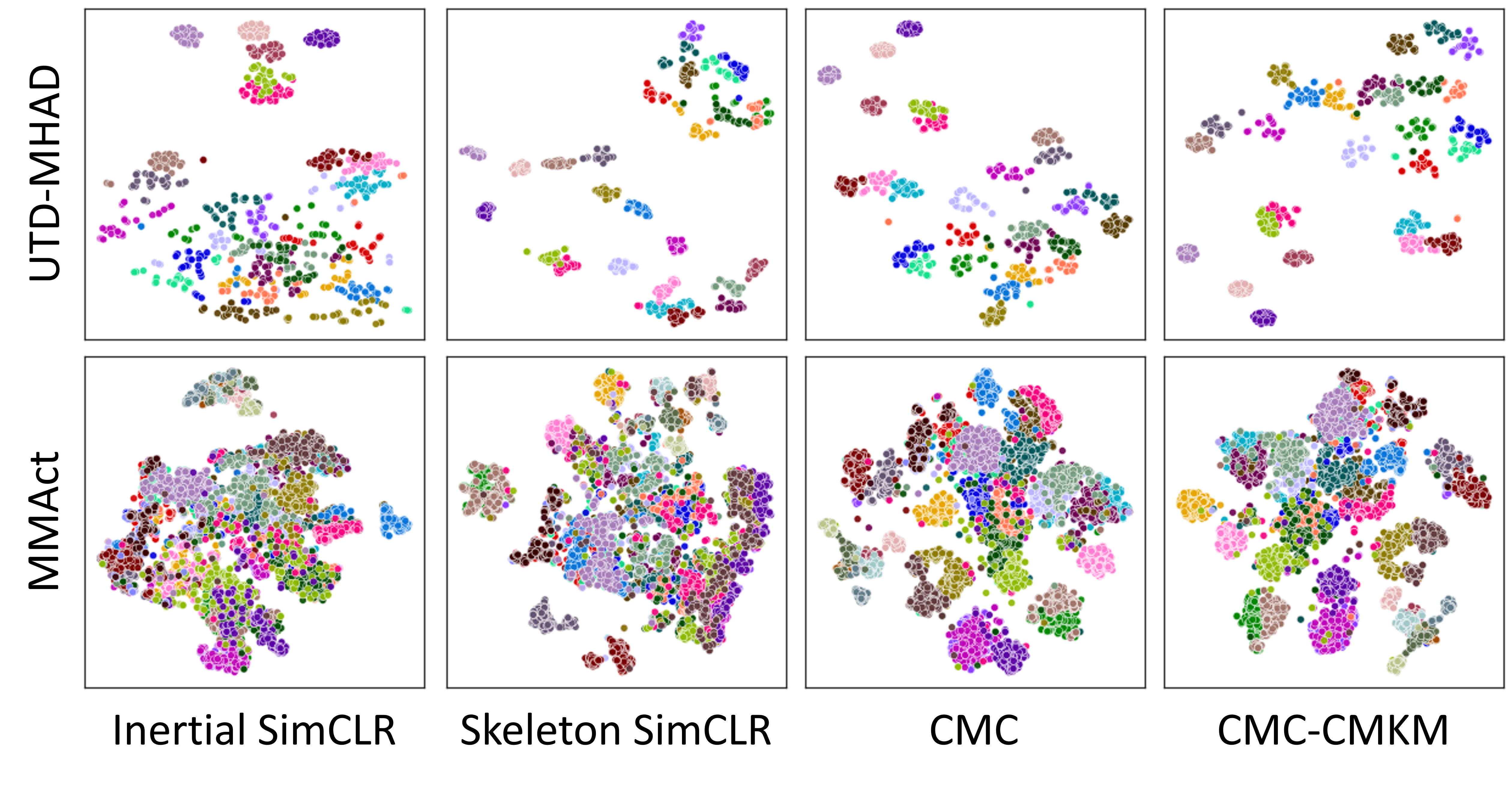}
}
\caption{Visualization of the learnt representations using t-SNE.}
\label{fig:tsne}
\vspace{-10pt}
\end{figure}

\noindent\textbf{Qualitative analysis of features.} In order to assess the separation between classes visually, we project the feature embeddings into a two-dimensional space using t-SNE \cite{Maaten_2008_tsne} and visualize them in Figure \ref{fig:tsne}. For the multimodal approach, we concatenate inertial and skeleton features before feeding them to the t-SNE. From the obtained diagrams, it is clear that features learnt on multimodal data contribute to better class separation.

\subsection{Semi-supervised Learning Scenario}

In a more realistic scenario, vast amounts of labeled training data might not be available. In this case, one can still use the unannotated dataset to pre-train models in the SSL manner. For semi-supervised learning evaluation, we limit the amount of labels available for training. Specifically, we perform an experiment where only a random percentage $p \in \{1\%, 2\%, 5\%, 10\%, 25\%, 50\%\}$ of labels is present. In this experiment, we compare the performance of the proposed multimodal SSL models to the supervised and random models. Namely, we pre-train CMC and CMC-CMKM using the whole unannotated dataset and fine-tune the fusion linear layers and the linear classifier with annotated data. Besides, we train a supervised multimodal model with identical encoders using these data. Finally, we also fine-tune the fusion network and the linear classifier for randomly initialized encoders. For each value of $p$ the experiment is repeated 10 times and the average performance values with the corresponding confidence intervals are presented in Figure \ref{fig:semi_sup}. We also include, as a horizontal dashed line, the performance of a fully supervised multimodal network ($p$=100\%).

According to the obtained figures, the multimodal SSL approaches are much more robust, especially when very limited amounts of annotated data are available. For both datasets, the the SSL models outperform the identical supervised models by more than 10\% when less than 10\% of data is annotated. What is more, the proposed SSL models almost reach the performance of the fully supervised model when more than 25\% of labeled data is available.
\section{Conclusions and Future Work}

In this paper, we adopted a set of modality-specific network architectures for encoding inertial and skeleton data and implemented modern unimodal and multimodal self-supervised frameworks, adapting them to the problem of HAR. Furthermore, we proposed a novel framework named CMC-CMKM, which addresses issues related to the CMC pre-training by injecting modality-specific knowledge into the learning process. Extensive experiments have shown that the proposed multimodal SSL frameworks outperform unimodal supervised approaches and show satisfactory performance comparing to multimodal fully-supervised models.

As for the future work, additional data modalities can be used in combination with different backbone architectures. Moreover, while the problems related to negative pairs have been mitigated to some extent using CMC-CMKM framework, there is an entire class of self-supervised approaches which does not rely on negative pairs and it might be useful to explore how they can be adapted to multimodal settings.

\bibliographystyle{IEEEtran}
\bibliography{bibliography}

\end{document}